\documentclass[conference]{IEEEtran}
\IEEEoverridecommandlockouts
\usepackage{cite}
\usepackage{amsmath,amssymb,amsfonts}
\usepackage{algorithmic}
\usepackage{multirow}
\usepackage{graphicx}
\usepackage{textcomp}
\usepackage{xcolor}
\usepackage{array}
\usepackage{pbox}
\usepackage{pifont}

\usepackage{fancyhdr}
\def\BibTeX{{\rm B\kern-.05em{\sc i\kern-.025em b}\kern-.08em
    T\kern-.1667em\lower.7ex\hbox{E}\kern-.125emX}}
\begin{document}

\newcommand{\Chat}[1]{\textcolor{orange}{[#1]}}
\newcommand{\ZY}[1]{\textcolor{purple}{[#1]}}
\newcommand{\Stella}[1]{\textcolor{blue}{[#1]}}

\title{Scaffolding Language Learning via Multi-modal Tutoring Systems with Pedagogical Instructions}

\author{
Zhengyuan Liu\textsuperscript{\ding{105}\ding{118}},\ Stella Xin Yin\textsuperscript{\ding{105}\ding{70}},\ Carolyn Lee\textsuperscript{\ding{171}},\ Nancy F. Chen\textsuperscript{\ding{118}}\\
\textsuperscript{\ding{70}}Nanyang Technological University, Singapore\ \ \ \textsuperscript{\ding{171}}Stanford University\\
\textsuperscript{\ding{118}}Institute for Infocomm Research (I$^2$R), A*STAR, Singapore\\
}

\maketitle

\begin{abstract}
Intelligent tutoring systems (ITSs) that imitate human tutors and aim to provide immediate and customized instructions or feedback to learners have shown their effectiveness in education. With the emergence of generative artificial intelligence, large language models (LLMs) further entitle the systems to complex and coherent conversational interactions. These systems would be of great help in language education as it involves developing skills in communication, which, however, drew relatively less attention. Additionally, due to the complicated cognitive development at younger ages, more endeavors are needed for practical uses. Scaffolding refers to a teaching technique where teachers provide support and guidance to students for learning and developing new concepts or skills. It is an effective way to support diverse learning needs, goals, processes, and outcomes. In this work, we investigate how pedagogical instructions facilitate the scaffolding in ITSs, by conducting a case study on guiding children to describe images for language learning. We construct different types of scaffolding tutoring systems grounded in four fundamental learning theories: knowledge construction, inquiry-based learning, dialogic teaching, and zone of proximal development. For qualitative and quantitative analyses, we build and refine a seven-dimension rubric to evaluate the scaffolding process. In our experiment on GPT-4V, we observe that LLMs demonstrate strong potential to follow pedagogical instructions and achieve self-paced learning in different student groups. Moreover, we extend our evaluation framework from a manual to an automated approach, paving the way to benchmark various conversational tutoring systems.
\end{abstract}

\begin{IEEEkeywords}
Intelligent Tutoring Systems, Scaffolding, Multi-modal Language Models
\end{IEEEkeywords}

\section{Introduction}
Intelligent Tutoring Systems (ITSs) are adaptive instructional systems equipped with Artificial Intelligence (AI) and Natural Language Processing (NLP) technologies and integrated educational methodologies \cite{Mousavinasab-2021}. These systems offer personalized learning content, instant feedback, and interactive learning experience to learners. A significant feature of ITSs is their ability to tailor instructional activities and strategies to align with the learners' different characteristics, experiences, and learning needs \cite{Keleş-2009}. Numerous studies have highlighted the effectiveness and broad applicability of ITSs across various educational fields \cite{Kulik-2016,Mousavinasab-2021}. 
On the other hand, with the emergence of generative artificial intelligence \cite{Wei-2022-emergent,Ouyang-2022-GPT}, large language models (LLMs) further empower interactive ITSs with exceptional capabilities on conversational interactions \cite{Chen-2023, Nye-2023}, and show great potential to support students learning outside of classrooms in various disciplines, and they can make online education personalized and more accessible \cite{Kasneci-2023-llmEDU}.

For the applications of ITSs, the majority of the studies were conducted within the field of computer science and mathematics education and primarily targeted for university students \cite{Mousavinasab-2021}. Compared to those fields which more focus on abstract concepts, learning a language involves developing skills in communication, including speaking, listening, reading, and writing, resulting in a higher interaction demand for ITSs. Given the diverse learning needs, the complexity of open-ended questions, and cognitive development at younger ages, it's more challenging for ITSs to be applied effectively in children's language education.

\begin{figure}[t!]
\centerline{\includegraphics[width=6.0cm]{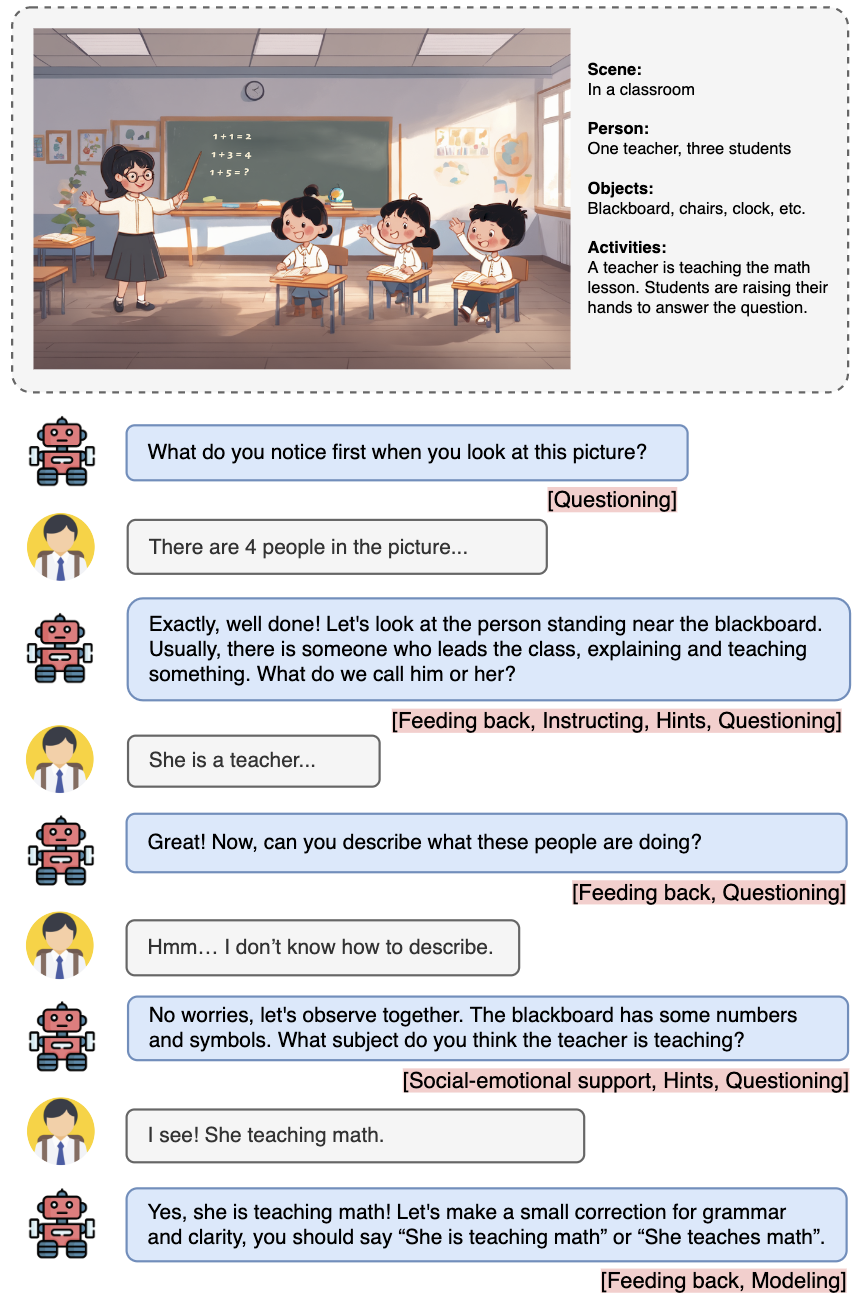}}
\caption{A dialogue example of interactive language learning via an image description tutoring system. The student is asked to describe the picture.}
\label{fig-intro-example}
\vspace{-0.4cm}
\end{figure}

\textit{``It is only when scaffolding is needed that learning will actually take place (Gibbons, 2002)''} 

To improve tutoring systems in language learning, a crucial aspect is providing effective scaffolding for learners. Scaffolding refers to an instructional technique in which teachers provide temporary and dynamic support and guidance to students as they learn and develop new concepts or skills \cite{Vygotsky-1978}. Over recent decades, scaffolding has been promoted as an effective way to support diverse learning needs, goals, processes, and outcomes. Its application is particularly effective in children’s language learning \cite{Gibbons-2002}. Previous studies have shown that learners are more likely to succeed in language learning when their teachers provide pedagogical support, facilitating a higher level of skill and understanding \cite{Hammond-2001,Gibbons-2002}.

In practice, teachers provide scaffolding by clarifying, questioning, and presenting models for the learners. They guide students through hints and suggestions, encouraging them to connect new information with their personal experiences and prior knowledge \cite{Gibbons-2002}. Since scaffolding is a dynamic intervention finely tuned to the learner’s ongoing progress, the support given by the teacher during scaffolding strongly depends on the patterns of teacher-student interactions \cite{van de Pol-2010}. Therefore, it would be of great help to improve the performance of ITSs by promoting the effectiveness of interactions.

In this work, we investigate how pedagogical instructions facilitate the scaffolding in LLM-based ITSs, and conduct a case study on guiding children to describe images for language learning. Following task-specific prompts, multi-modal LLMs can effectively organize the conversation by asking proper questions, giving constructive suggestions, and providing informative hints.
We build different types of conversational ITS grounded in four fundamental learning theories: knowledge construction, inquiry-based learning, dialogic teaching, and zone of proximal development, and compare them in simulated teaching sessions of a set of images and two student groups. To evaluate the scaffolding process of conversational interactions, we build and refine a seven-dimension rubric, which is employed in both qualitative and quantitative evaluations through feedback, hints, instruction, explaining, modeling, questioning, and social-emotional support. In our experiments on GPT-4V, we observe that LLMs can demonstrate strong potential to follow pedagogical instructions and achieve self-paced learning in different student groups. Furthermore, we transform our evaluation framework from manual to automated by leveraging the in-context learning capability of LLMs, and this paves the way to benchmark various tutoring systems.

\section{Related Work}
\subsection{Intelligent Tutoring Systems}
Intelligent tutoring systems (ITSs) aim to provide personalized and effective instructional support to students, and they have gained increasing importance due to the growing demand for adaptive and accessible education, especially in remote and online learning environments.
One significant approach to developing ITSs is to leverage various statistical features to perform learning analytics activities and performance prediction \cite{Weng-2020-analytics,Ouyang-2023-analytics}, and many previous studies have focused on engagement and dropout prevention, such as leveraging students’ facial expressions for emotion recognition and engagement prediction \cite{Leony-2013-emotions}.
On the other hand, as an advanced form of ITSs, conversational ITSs have been extensively investigated as educational dialogue systems \cite{Macina-2023-tutor,Ruan-2019-quizbot,Wollny-2021}, as they can provide adaptive instructions and real-time feedback to students. Most existing studies focus on learning the pedagogical strategies to teach the students of the given exercises \cite{Stasaski-2020,Suresh-2022}, or generating high-quality responses in the tutoring dialogues \cite{Wang-2023-Strategize}. Latest studies \cite{Chen-2023-empower,Dan-2023-eduChat,Nye-2023} on interactive ITSs powered by LLMs have showcased the exceptional capabilities of natural language interactions.

\subsection{Scaffolding in Children’s Language Learning}
Since the late 1970s, scaffolding has gained increasing popularity across various educational fields, especially in language learning contexts. This popularity is due to the crucial roles of the meaning-making process and linguistic assistance in students’ language development \cite{Walqui-2006, Kayi-Aydar-2013}. Teachers can apply scaffolding strategies, such as questioning, reformulation, repetition, and elaboration to assist English language learners in co-constructing content knowledge, thereby making these processes “visible” to them \cite{Gibbons-2002}. With the support and guidance of teachers, students are more likely to complete the given task and face similar challenges in the future with greater confidence \cite{de Oliveira-2023, Damanhouri-2021}. 
Previous research has identified several key characteristics essential for effective scaffolding \cite{Gonulal-2018}. The most salient feature is contingency. Teachers assess students' competency levels and dynamically adapt scaffolding strategies based on the learners' understanding and actions \cite{Walqui-2006}. Another aspect of scaffolding is fading \cite{Puntambekar-2005, Puntambekar-2022}. In this process, teachers gradually withdraw the scaffolding as students are able to carry out tasks independently \cite{Lajoie-2005}. Thus, scaffolding is a temporary and adjustable process, with support aligned towards facilitating students’ learning goals.

While research on scaffolding has enriched our understanding of language teaching practices, the scaffolding process is often limited to either one-on-one or one-to-many teacher-led instruction. This can result in limited access and fewer opportunities for students to engage in practice. Consequently, students might have fewer chances of being heard, scaffolded, and receiving feedback. Such limitations could adversely affect their language learning and usage. The advent of LLMs has empowered ITSs to provide scaffolding strategies in supporting students learning outside of classrooms in various disciplines, and they can make online learning personalized and more accessible.

\begin{table*}[t!]
\caption{Table of the description of pedagogical instructions.}
\begin{center}
\begin{tabular}{|p{3.0cm}|p{6.0cm}|p{7.7cm}|}
\hline
\multicolumn{1}{|c|}{\textbf{Theory Type}} & \multicolumn{1}{c|}{\textbf{Definition}} & \multicolumn{1}{c|}{\textbf{Pedagogical Strategy}} \\ \hline
Knowledge Construction\newline (Sullivan Palincsar, 1998) & The effortful, situated, and reflective process by which students solve problems and construct an understanding of concepts, phenomena, and situations. & Consistently assisting students in building upon their prior knowledge, organizing and synthesizing information, integrating ideas, and making inferences. \\ \hline
Inquiry-based Learning\newline (Pedaste et al., 2015) & Engaging learners by creating real-world connections through questioning and exploration. & Guiding learners with explicit learning goals and helping them develop an explanatory learning process, breaking down complex tasks into small and manageable segments, making observations, asking questions, posing hypotheses, investigating, interpreting, and discussing. \\ \hline
Dialogic Teaching\newline (Alexander, 2006) & The ongoing process of dialogue in stimulating and developing students' thinking, learning and understanding. & Co-constructing knowledge through dialogue and collaboration, encouraging the free exchange of ideas, using follow-up questions, clues, elaborations, reformulations, confirmations, or recaps, building on prior knowledge and understanding. \\ \hline
Zone of Proximal Development (Vygotsky, 1978) & The space between what a learner can do without assistance and what a learner can do with adult guidance or in collaboration with more capable peers. & Assessing the learner’s current ability level, connecting content to learners’ existing knowledge, breaking down a task into smaller, manageable components, and using prompts and cues to help students achieve a potential level beyond their current capabilities. \\ \hline
\end{tabular}
\label{table-instructions-def}
\end{center}
\vspace{-0.2cm}
\end{table*}

\section{Multi-modal Tutoring Systems with Pedagogical Instructions}

\subsection{Tutoring Language Learning via Image Description}
Teaching and improving primary students' language learning through image description is a dynamic and engaging approach \cite{Wright-1989}. As the example shown in Figure \ref{fig-intro-example}, a learning session usually begins by presenting an image and encouraging students to observe it closely, then the teacher asks open-ended questions to stimulate their thinking, such as ``\textit{What do you see happening in this picture?}'' or ``\textit{Can you describe the people or animals you see?}''

Beyond merely listing the objects in the image, the teacher further guides students to describe how things look, feel, or sound, and encourages students to use adjectives and adverbs. This exercise enhances their language skills including vocabulary, organization, and fluency \cite{Pinter-2017}. To further develop their language skills, the teacher introduces new vocabulary related to the image, and engages students in storytelling. Students are asked to create a short story or summary, thereby stimulating their creativity and imagination to enrich the narrative with additional details and depth.

\subsection{Multi-modal Systems as an Image Description Tutor}
Beyond text-based interactive learning (e.g., math and coding tutoring), multi-modal capabilities are essential for building image description tutoring systems. Basically, it includes four functional aspects: vision modeling, speech recognition, natural language generation, and dialogue management. More specifically, given an image input, vision modeling is to capture the visual features and recognize various scenes, objects, and activities, as well as grounded knowledge such as spatial relationships \cite{Yamada-2023}. Speech recognition is to convert student responses from audio to text. Natural language generation and dialogue management enable the tutoring system to interact with students via effective communication, including generating fluent, coherent, and descriptive language of the image, raising contextualized questions, and providing hints and explanations. In oral courses, spoken language assessment is also an integral component within machine-aided language learning, which is used to evaluate oral proficiency \cite{Wong-2022-spoken}.

Upon the versatility and capability of LLMs, one can build tutoring systems without massive supervision from time-consuming manual annotation \cite{Stasaski-2020,Suresh-2022}. Therefore, we leverage GPT-4V as an image description tutoring agent for language learning, since it is a multi-modal model that supports all four functional aspects, and shows state-of-the-art instruction-following and reasoning capabilities \cite{Yang-2023-gpt4v-eval}.

\begin{table*}[t!]
\caption{Table of the rubric definition for evaluating scaffolding efficacy.}
\begin{center}
\begin{tabular}{|cl|p{6.3cm}|p{5.5cm}|}
\hline
\multicolumn{2}{|c|}{\textbf{Dimension}} & \multicolumn{1}{c|}{\textbf{Definition}} & \multicolumn{1}{c|}{\textbf{Utterance Example}} \\ \hline
\multicolumn{1}{|c|}{\multirow[t]{6}{*}{Cognitive Scaffolding}} & Feeding back & The teacher directly evaluates the behavior or response of the student. & Yes, the girl does look happy! \newline Great! You're right. \\ \cline{2-4} 
\multicolumn{1}{|c|}{} & Hints & The teacher gives an explicit hint with respect to the expected answer. & Does he look happy, surprised, or something else? Look at the TV in the picture for a clue. \\ \cline{2-4}
\multicolumn{1}{|c|}{} & Instructing & The teacher provides information so that the student knows what to do or how to do it. Request for a specific action (e.g., look at sth. or focus sth.). & Look at the things around them for clues.\newline Remember to include what you've noticed about cleaning and organizing. \\ \cline{2-4} 
\multicolumn{1}{|c|}{} & Explaining & The teacher provides detailed information on ``why'' or clarification. & When someone opens their mouth like that and has tears on their face, it often does indicate that they are crying or upset. \\ \cline{2-4} 
\multicolumn{1}{|c|}{} & Modeling & The teacher demonstrates behavior (verbal or non-verbal) for imitation. & Just a small grammar tip: when we say ``with the girl is dancing,'' we don't need the word ``is'' after ``girl''. \\ \cline{2-4}
\multicolumn{1}{|c|}{} & Questioning & The teacher asks the student questions that require an active linguistic and cognitive answer. & Can you tell me if it's daytime or nighttime? And how can you tell? \\ \hline
\multicolumn{1}{|l|}{Social-emotional Support} &  & Responses related to emotion and motivation such as positive affirmation, showing empathy, promoting self-efficacy, fostering a sense of connectedness, encouraging perseverance, and other related constructs. & No problem at all!\newline No worries, let's observe together!\\ \hline
\end{tabular}
\label{table-rubric-definition}
\end{center}
\vspace{-0.3cm}
\end{table*}

\subsection{Enhancing Tutoring Systems with Pedagogical Instructions}
In practical settings, teachers adhere to established pedagogical principles to enhance their instructional methods, demonstrating the efficacy of a more focused and systematic approach \cite{Wells-1999}. Consequently, to develop an image-based tutoring system that effectively motivates and supports students in language acquisition, we explore the impact of incorporating structured pedagogical strategies.
On the other hand, LLMs are capable of following complex and detailed prompts, and performing as task-specialized agents \cite{Ouyang-2022}.
Previous work shows that prompting in a structured manner is beneficial for complex instruction following \cite{Khot-2022-Prompt}, thus we split it into three parts: role \& task definition, pedagogical instruction, and behavior constraint. Here is one template:\\
\noindent \textit{\textbf{[Role \& Task Definition]} You are a primary school language teacher who teaches me to describe the picture.}\\
\noindent \textit{\textbf{[Pedagogical Instruction]} You are using the knowledge construction approach. This involves any one of the following: building on prior knowledge, selecting information, organizing information, integrating ideas, making inferences, and helping me describe the picture.}\\
\noindent \textit{\textbf{[Behavior Constraint]} Ask me only one question at a time. Always wait for my input before proceeding to the next step. Correct my answers if they are inaccurate.}

\noindent Moreover, for the pedagogical instruction, we apply four constructivist learning theories (as shown in Table \ref{table-instructions-def}) and conduct experiments on how they affect the scaffolding of language learning via image description.

\noindent \textbf{Knowledge Construction} Constructivist researchers believe that when individuals encounter new information, they rely on their prior knowledge and personal experience to interpret it \cite{Resnick-1987}. During this meaning-making process, individuals reformulate the new information or restructure their existing knowledge, thereby achieving deeper understanding \cite{Palincsar-1998}. Prior research found that knowledge construction can range from simple restatements and paraphrasing to more complex activities like explanations, inferences, justifications, hypotheses, and speculations \cite{Chan-1992}. Specifically, teachers facilitate students' knowledge construction process by consistently assisting students in building upon their prior knowledge, organizing and synthesizing information, integrating ideas, and making inferences \cite{van Aalst-2009}.

\noindent \textbf{Inquiry-based Learning} is a pedagogical approach that engages learners by creating real-world connections through questioning and exploration \cite{Pedaste-2015}. It aims to inspire students to take ownership of their learning journey. To support these inquiry outcomes, researchers have proposed several scaffolding strategies \cite{McNeill-2006,Reiser-2004}. For example, teachers guide learners with explicit learning goals and help them develop an explanatory learning process \cite{Hsu-2015}. Specifically, tasks are structured to minimize cognitive overload. Teachers break down complex tasks into small and manageable segments. This approach narrows the “problem space”, enabling learners to focus their efforts and utilize available resources or tools effectively \cite{Reiser-2004}.

\noindent \textbf{Dialogic Teaching} highlights the role of talk in stimulating and developing students' thinking, learning, and understanding \cite{Alexander-2000,Alexander-2006}. To facilitate productive interactions, teachers usually encourage students by posing thought-provoking questions and inviting them to share their knowledge and experiences, which aims not just to seek the right answers, but also to elicit reasons and explanations \cite{Mercer-2012}. This often involves a scaffolded dialogue pattern known as initiation-response-feedback pattern (IRF) \cite{Wells-1999}. That is, the teacher initiates a topic, students respond, and then the teacher provides feedback \cite{Wells-2006}. In this cyclic IRF sequence, teachers guide students by using follow-up questions, clues, elaborations, reformulations, confirmations, or recaps, thereby maintaining students’ active participation \cite{Alexander-2018}.

\noindent \textbf{Zone of Proximal Development (ZPD)}  is defined as the space between what a learner can do without assistance and what a learner can do with adult guidance or in collaboration with more capable peers \cite{Vygotsky-1978}. In Vygotsky’s view, a learner’s ability to bridge this gap between actual performance and potential ability depends on the scaffolding provided by more capable others \cite{Raymond-2000}. In pedagogical contexts, scaffolding techniques involve several processes: assessing the learner's current ability level, connecting content to learners' existing knowledge, breaking down a task into smaller, manageable components, and using prompts and cues to help students achieve a potential level beyond their current capabilities. 

Based on the aforementioned pedagogical theories, we summarized key features of each theory and synthesized corresponding instructional strategies for scaffolding within our ITS, as shown in Table \ref{table-instructions-def}. These strategies, embedded in student-ITS conversational interactions, aim to provide guidance and support in image description tasks, thereby enhancing the overall learning experience.

\begin{figure*}[t!]
\centerline{\includegraphics[width=14.0cm]{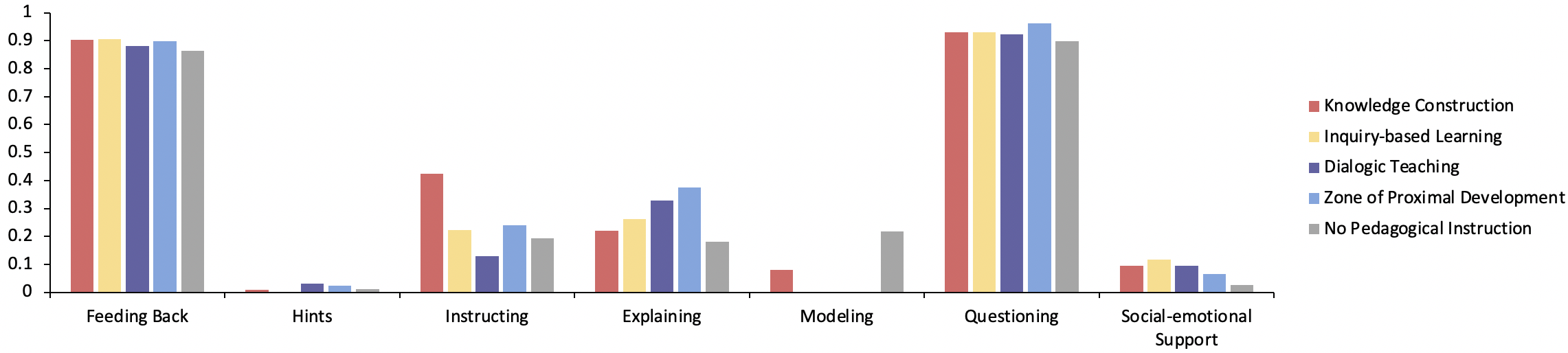}}
\centerline{\includegraphics[width=14.0cm]{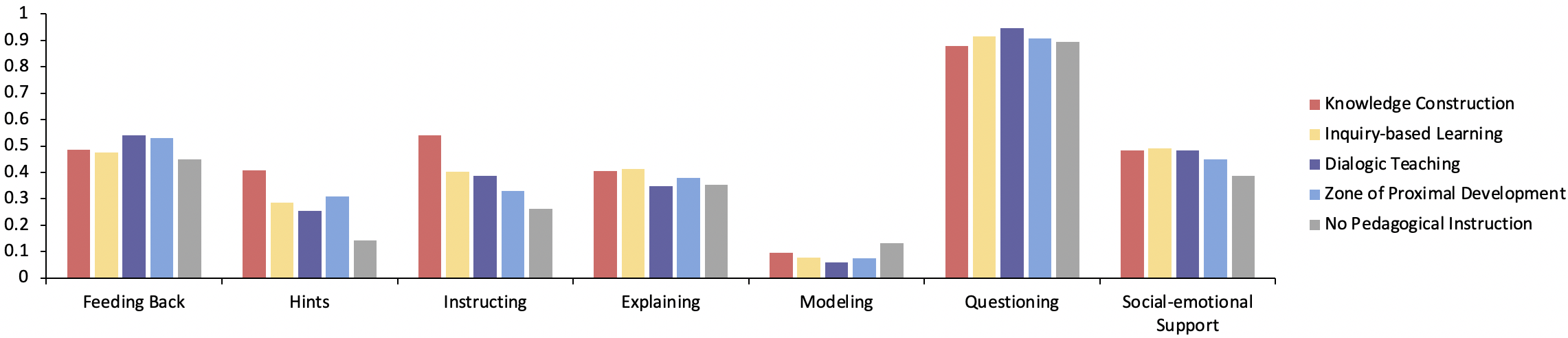}}
\vspace{-0.2cm}
\caption{Coding result of the five systems with different pedagogical instructions (Up: high-ability group; Down: low-ability group).}
\label{fig-average-comparison}
\vspace{-0.1cm}
\end{figure*}

\begin{figure*}[t!]
\centerline{\includegraphics[width=18.1cm]{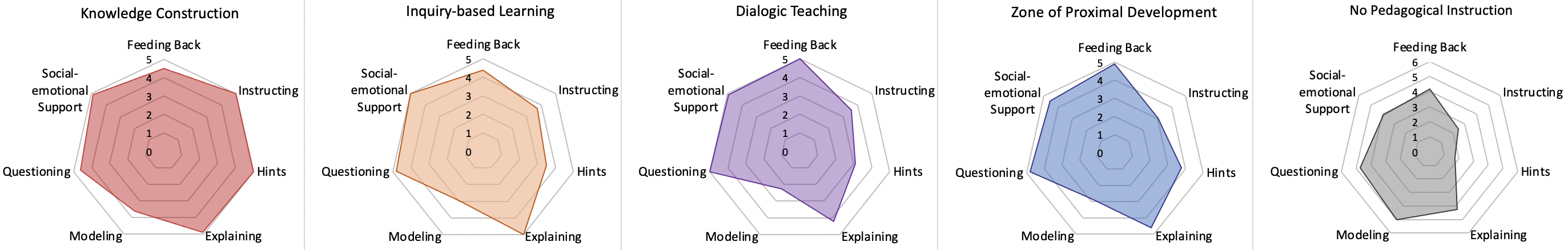}}
\caption{Normalized capability scoring in seven dimensions of the five systems with different pedagogical instructions.}
\vspace{-0.2cm}
\label{fig-radar-comparison}
\vspace{-0.2cm}
\end{figure*}

\section{Evaluating Tutoring Systems from Scaffolding Perspective}

\subsection{Building Student Capability Levels for Evaluation}
Scaffolding strategies effectively support the learning processes of students. However, the needs, learning styles, and educational experiences of low- and high-achieving learners differ significantly \cite{Hargis-2006}. First, low achievers often feel uncomfortable expressing their ideas because they may lack prior knowledge and self-confidence. They tend to wait for assistance rather than attempting to solve problems independently \cite{Soller-1998,Haruehansawasin-2018}. Second, students with lower performance frequently encounter more misconceptions resulting in the need for more individualized learning paths and more interactive and adaptive scaffolds \cite{Reinhold-2020,Sweller-2011}.

These differences have led to the classification of students into \textit{high-} and \textit{low-ability} groups. In this study, \textit{high-ability} students are defined as those with high language proficiency who can answer each question correctly with minimal support and guidance. Conversely, \textit{low-ability} students are characterized by low language proficiency and are not able to answer questions independently. This classification serves two main purposes: first, to investigate how scaffolding strategies are applied to students with varying abilities, and second, to explore the differences among different tutoring systems.

\subsection{Building Scaffolding Evaluation Rubric}
When comparing the learning efficacy of tutoring systems, it is common to conduct assessments on post-learning performance, dropout rate, and user feedback. However, the scaffolding strategy in conversational interactions, another important aspect, is overlooked \cite{Duffy-2015}.
To evaluate the effectiveness of scaffolding strategies, here we introduce and refine a rubric of dialogic analysis at the utterance level. Based on previous pedagogical work \cite{Wells-1999,van de Pol-2010}, our rubric is designed to understand how scaffolding is performed during students' language learning, and it consists of seven dimensions (as shown in Table \ref{table-rubric-definition}): Feeding back, Hints, Instructing, Explaining, Modeling, Questioning, and Social-emotional Support.

\subsection{Experimental Setting and Qualitative Analysis}
Inspired by real-world learning materials for primary school level 1 and level 2 students, we constructed a seed dataset for qualitative analysis. To improve the diversity of visual and language features, the selected 10 image description samples cover various scenes (e.g., classroom, playground, home), objects (e.g., family, kids, teacher), and activities (e.g., sports, reading).
We simulate the learning process via human-machine interaction, where the tutoring system leads the conversation, and we feed user responses according to the assigned student group. The average turn number of each conversation is 22.5, we repeat each session across 5 pedagogical instruction types as well as 2 student groups (a system without pedagogical instruction is added as control), and the total collected utterance number is 2250.
In our preliminary analysis, we observe that the system without any pedagogical instruction is also capable of utilizing visual features and organizing the conversation. For instance, when students get confused, the tutor will ask them to look at one specific part of the picture and encourage their attention to certain objects.

Based on our rubric, we code the system utterances and calculate the scores among five types of pedagogical instructions and between two students’ ability levels. For each system utterance, a score of 1 was assigned if it corresponded with the dimensions of scaffolding strategies. Two linguistic annotators coded system generations independently, and we conducted two rounds of preliminary coding to consolidate the rubric description and reduce discrepancies. The final inter-annotator Cohen's Kappa is 0.75.

\begin{table*}[t!]
\linespread{1.1}
\caption{Experimental results of automated scoring by leveraging LLMs as evaluator.}
\begin{center}
\begin{tabular}{|p{3.0cm}|p{1.7cm}<{\centering}|p{1.7cm}<{\centering}|p{1.7cm}<{\centering}|p{1.7cm}<{\centering}|p{1.7cm}<{\centering}|p{1.7cm}<{\centering}|p{1.7cm}<{\centering}|}
\hline
 & \multicolumn{2}{c|}{\textbf{Zero-Shot Inference}} & \multicolumn{2}{c|}{\textbf{1-Shot Inference}} & \multicolumn{2}{c|}{\textbf{3-Shot Inference}} \\
 \hline
\textbf{Model} & \textbf{Accuracy} & \textbf{F1 Score} & \textbf{Accuracy} & \textbf{F1 Score} & \textbf{Accuracy} & \textbf{F1 Score} \\ \hline
LLaMA-2-7B-Chat & 0.533 & 0.497 & 0.548 & 0.536 & 0.654 & 0.644 \\ \hline
LLaMA-2-13B-Chat & 0.599 & 0.375 & 0.609 & 0.583 & 0.708 & 0.698 \\ \hline
Vicuna-13B-V1.5 & 0.417 & 0.368 & 0.670 & 0.650 & 0.765 & 0.757 \\ \hline
Mistral-7B-Instruct-V0.1 & 0.480 & 0.426 & 0.743 & 0.733 & 0.777 & 0.769 \\ \hline
Zephyr-7B-Beta & 0.711 & 0.706 & 0.746 & 0.732 & 0.785 & 0.778 \\ \hline
GPT-3.5-turbo-1106 & 0.783 & 0.764 & 0.787 & 0.775 & 0.805 & 0.795 \\ \hline
\end{tabular}
\label{table-LLM-eval-result}
\end{center}
\vspace{-0.4cm}
\end{table*}

\subsection{Experimental Results and Analysis}

\subsubsection{Comparison between high- and low-ability students}
The concept of contingency emphasized the malleable feature of scaffolding in relation to students' understanding. Contingent support suggests that the tutor amplifies the level of support in reaction to student failure or diminishes it following student success. In this study, we compared the scaffolding strategies that ITS applied to \textit{high-} and \textit{low- ability} students. We observed that systems with four pedagogical instructions outperformed in providing contingent support, managing to increase the degree of contingency for \textit{low- ability} students while reducing it for those with high abilities. This increased level of contingency was associated with a rise in the provision of hints, instruction, explanations, modeling, and social-emotional support.

Specifically, in Figure \ref{fig-average-comparison}, we observed that when engaging with \textit{high-ability} students, the scaffolding typically begins with positive affirmation, followed by guiding the students through questions and clarifying their answers. In contrast, when interacting with \textit{low-ability} students, systems tend to pay more attention to social-emotional support. Additionally, the scaffolding for these students includes more hints and explanations, and in most cases, they provide structured examples for the students to imitate.

\subsubsection{Comparison among Scaffolding Tutors}
Figure \ref{fig-radar-comparison} illustrates capability scores in 7 dimensions of various scaffolding strategies (normalized by the max value of each dimension). Those equipped with pedagogical instructions outperformed the ones without pedagogical instruction in each dimension except for \textit{Modeling}. This can be attributed to the difference of \textit{Modeling} contents. The system lacking pedagogical instructions predominantly offers direct answers to students who are unable to answer questions themselves. In contrast, other ones, particularly when interacting with students of lower language proficiency, initiate with hints and explanations, aiming to encourage and assist the students.

Hints are one of the most supportive dimensions in facilitating students with language learning \cite{Celce-Murcia-2000}. During the scaffolding process, hints work as moderators between students and knowledge. They effectively help learners access contextual information and important shortcuts, ultimately assisting students in language development \cite{Khaliliaqdam-2014}. In this study, the occurrence of hints is significantly more prevalent among pedagogical-based tutoring systems, particularly in interactions with \textit{low-ability} students due to their limited vocabulary and lower language proficiency. This suggests that they are capable of dynamically adapting their scaffolding approach to meet the diverse needs of learners, prioritizing language support where it is most needed.

Language learning occurs through imitation, reinforcement of contextual or verbal stimuli, practice of correct responses, and immediate correction of incorrect responses by the teacher \cite{Dulay-1973}. Additionally, several strategies have been identified for introducing or explaining new topics, concepts, or terms to students, including explaining, reformulating, clarifying, and exemplifying \cite{Richards-2014}. These approaches effectively build connections between new information and students' prior knowledge. In this study, we observed that tutors with pedagogical instructions are capable of explaining, describing, elaborating, and comparing new knowledge by leveraging students’ familiar concepts. They could also provide grammatical structures, contextual forms, or examples for correction, thus enabling students to construct descriptions accurately. Conversely, the tutor without pedagogical instructions often provides direct answers with less personalized scaffolding and fewer interactive learning opportunities, leading to a more passive and less engaging learning experience for students.

\section{Towards Automated Scaffolding Evaluation}

\subsection{Leveraging LLMs for Automated Evaluation}
Since the student scaffolding can be significantly affected by the pedagogical tutor, in practical use cases, transforming from manual to automated utterance scoring represents a significant advancement in scalable, accurate, and efficient evaluation. In this section, we investigate the potential of employing LLMs for automated scoring. Recent work shows that LLMs can achieve a high correlation with human judgments on various tasks \cite{Li-2023-Annotator}. Based on previous work, we designed a natural language instruction for the scoring task according to our rubric. The prompt is created by concatenating the scoring criteria, and utterance context, and then fed to a model for prediction. The output is the labeled types for each dimension based on the defined schema.

\subsection{Experimental Results and Analysis}
To demonstrate the feasibility and efficacy of automated evaluation and compare the performance of LLMs, we use our manual annotation as a reference, and results are presented in terms of correlation with human judgments, using accuracy and F1 scores.
Here we selected and tested a list of representative models. As shown in Table \ref{table-LLM-eval-result}, while most models cannot provide reasonable results under zero-shot inference setting, the performance can be significantly improved by adding only 3 examples (i.e., 3-shot inference). In particular, for LLaMA-based models (i.e., LLaMA-2 and Vicuna), a larger parameter size (13B vs. 7B) brings higher accuracy. The 7B models can achieve state-of-the-art performance (e.g., Mistral-7B, Zephyr-7B), and are comparable to GPT-3.5 in the few-shot setting. This demonstrates the feasibility of utilizing open LLMs to build automated and scalable scaffolding benchmarks.

\section{Conclusion}
Our work contributes to an in-depth understanding of LLM-based ITSs from the scaffolding perspective. First, we built a tutoring system that guides children to describe images for language learning, and enhanced it with pedagogical instructions. Second, we developed and validated a seven-dimension rubric to assess scaffolding strategies for different student groups. Our findings offer valuable insights into instructional design, improving the learning experience through interactive and supportive scaffolding strategies, which are aligned with personalized learning needs. Third, we leveraged LLMs to automate the scaffolding evaluation framework, paving the way to benchmark various conversational tutoring systems.


\end{document}